*Article*

# A machine learning-based severity prediction tool for diabetic sensorimotor polyneuropathy using Michigan neuropathy screening instrumentations

Fahmida Haque [1], Mamun B. I. Reaz [1*], Muhammad E. H. Chowdhury[2*], Rayaz Malik[3], Mohammed Alhatou[4], Syoji Kobashi[5], Iffat Ara[2], Sawal H. M. Ali[1], Ahmad A. A Bakar[1], Geetika Srivastava[6],

[1] Department of Electrical, Electronic and System Engineering, Universiti Kebangsaan Malaysia, Bangi, 43600, Selangor, Malaysia; email: F.H. (fahmida32@yahoo.com), M.B.I.R. (mamun@ukm.edu.my), S,H.M.A. (sawal@ukm.edu.my), A.A.A.B (ashrif@ukm.edu.my)

[2] Department of Electrical Engineering, Qatar University, Doha, 2713, Qatar; email: M.E.H.C. (mchowdhury@qu.edu.qa), I.A. (iffat.ara@qu.edu.qa)

[3] Weill Cornell Medicine – Qatar, Doha, Qatar; email: ram2045@qatar-med.cornell.edu

[4] Neuromuscular Division, Hamad General Hospital and Department of Neurology. Alkhor Hospital, Doha, 3050, Qatar; email: alhatou@yahoo.com

[5] Graduate School of Engineering, University of Hyogo, Hyogo, Japan; email: kobashi@ieee.org

[6] Department of Physics and Electronics, Dr Ram Manohar Lohia Avadh University, Ayodhya, 224001, India. Email: GS (gsrivastava@rmlau.ac.in)

*Corresponding author: mchowdhury@qu.edu.qa; mamun@ukm.edu.my

**Abstract:**

**Background:** Diabetic Sensorimotor polyneuropathy (DSPN) is a major long-term complication in diabetic patients associated with painful neuropathy, foot ulceration and amputation. The Michigan neuropathy screening instrument (MNSI) is one of the most common screening techniques for DSPN, however, it does not provide any direct severity grading system.

**Method:** For designing and modelling the DSPN severity grading systems for MNSI, 19 years of data from Epidemiology of Diabetes Interventions and Complications (EDIC) clinical trials were used. MNSI variables and patient outcomes were investigated using machine learning tools to identify the features having higher association in DSPN identification. A multivariable logistic regression-based nomogram was generated and validated for DSPN severity grading.

**Results:** The top-7 ranked features from MNSI: 10-gm filament, Vibration perception (R), Vibration perception (L), previous diabetic neuropathy, the appearance of deformities, appearance of callus and appearance of fissure were identified as key features for identifying DSPN using the extra tree model. The area under the curve (AUC) of the nomogram for the internal and external datasets were 0.9421 and 0.946, respectively. From the developed nomogram, the probability of having DSPN was predicted and a DSPN severity scoring system for MNSI was developed from the probability score. The model performance was validated on an independent dataset. Patients were stratified into four severity levels: absent, mild, moderate, and severe using a cut-off value of 10.5, 12.7 and 15 for a DSPN probability less than 50%, 75% to 90%, and above 90%, respectively.

**Conclusions:** This study provides a simple, easy-to-use and reliable algorithm for defining the prognosis and management of patients with DSPN.

## 1. Introduction

Diabetic sensorimotor polyneuropathy (DSPN) is the commonest long-term complication of diabetes and in the long term leads to ulceration and amputation and increased mortality [1]. Early identification is key to improving risk factors to prevent the progression of DSPN [2–5]. The American Diabetic Association (ADA) [1] and Toronto [6] consensus statements recommended that the diagnosis of DSPN should be based on symptoms and signs and nerve conduction studies (NCS). Many diagnostic techniques are available for DSPN [1,7–13] alongside several composite scoring techniques for severity stratification [14–17]. Toronto consensus endorsed the use of composite screening techniques for DSPN severity identification [6].

The Michigan neuropathy screening instrument (MNSI), commonly used composite scoring techniques, is recommended for the clinical diagnosis of DSPN on the ADA position statement [1]. The MNSI is a simple, inexpensive, reliable, and accurate assessment [17–19] that has been used to identify DSPN in many studies and clinical trials [18,20–25]. Neuropathy symptoms are assessed from 15 yes/no questions and neuropathy signs are assessed from five simple clinical tests. Patient is considered DSPN if the total score is ≥ 7 or ≥ 2 on MNSI questionnaire or examination respectively [17]. There is controversy on the optimal cut-off value for identifying DSPN with studies suggesting different cut-offs ranging from 2 to 1.5 [26], 2.5 [26–28], 3 [26] and 4 [29]. Moghtaderi et al. [26] reported an MNSI cut-off of 2 with a reliability of 0.81. Other studies have reported 80% sensitivity and 95% specificity and good repeatability for MNSI examination cut-off ≥ 2.0 [17,30]. Herman et al. [27] suggested the use of MNSI in larger clinical trials due to its easily applicable and non-invasive clinical tests compared to NCS. However, MNSI doesn't have any standardized grading system for severity classification.

Recently, machine learning (ML) methods have been used successfully to solve different disease prediction and classification problems [31–33] because of their ability and reliability in extracting information from complex, non-linear, or incomplete data supporting healthcare professionals with decision-making [34–38]. The fuzzy inference system (FIS) [39–42], multi-category support vector machine (SVM) learning [43], and adaptive fuzzy inference system (ANFIS) [44] have been reported to aid in the identification and stratification of diabetic neuropathy (DN). However, classifiers using fuzzy systems do not appear to be reliable as they rely on the if-then rule base set. Kazemi et al. [43], developed a multiclass SVM based DSPN severity classifier using the neuropathy disability score (NDS) and reported an accuracy of 76%. We have reported [44] an accuracy of 91% for an ANFIS for DSPN severity classification using MNSI utilizing 3 MNSI variables (questionnaire, vibration perception, and tactile sensitivity). Reddy et al. [45] identified various risk factors for DN and proposed a Radial basis function (RBF) network for DN prediction which only achieved 68.18% accuracy. Chen et al. [46] developed a prediction model to identify diabetic peripheral neuropathy (DPN) using logistic regression (LR) and reported the value of the concordance index (c-index) of 0.75 using MNSI.

In this research, with the help of ML a DSPN severity grading system will be developed for MNSI data. Initially, the most appropriate MNSI features were identified, based on which a multivariable logistic regression-based nomogram was developed and validated for the severity classification of the DSPN patients.

## 2. Materials and Methods

*2.1 Database Description*

For this research, two different Michigan Neuropathy Screening Instrument datasets have been collected. The first dataset was collected from the Epidemiology of Diabetes Interventions and Complications (EDIC) clinical trials [47,48]. IN EDIC trials, MNSI was used for annual assessment of DSPN in the recruited type 1 diabetes patients [47,48]. A detailed description of the EDIC trials procedures and baseline characteristics of the patients have been described previously [47–51].

For validation of our model, the second MNSI dataset was made available to us by Watari et al. [42] comprised of 102 patients with 21 MNSI variables: 15 questionnaires, Vibration perception (R), Vibration perception (L), 10-gm filament (combined results from both legs), Appearance of Deformities (combined results from both legs), Appearance of Callus (combined results from both legs), Appearance of Fissure (combined results from both legs). To have consistency for both datasets, we have considered 21 variables to design our prediction model.

*2.2 Data Imputation*

In practice, missing values in clinical data from larger clinical trials like EDIC is a quite common phenomenon. As the training of ML models highly depends on the dataset provided, missing data can be misleading for ML model training. To overcome this issue, data imputation techniques are applied [52,53]. 19 years MNSI data from EDIC trials with 14166 samples were collected. However, there were many duplicate responses which was removed, and 3754 unique samples were retrieved. In this study, missing data were imputed by the Multiple Imputation by Chained Equations (MICE) technique [52–55].

*2.3 Feature Ranking*

To identify the best combinations of MNSI features to identify DSPN, three different feature ranking techniques: Multi-Tree Extreme Gradient Boost (XGBoost) [56,57], Random Forest (RF) [58], Extremely Randomized Trees (Extra Tree) [59] techniques have been studied, and the best performing algorithm was reported. In-house code was written using Python 3.7 environment for data imputation and feature ranking.

*2.4 Logistic Regression Classifiers*

A supervised logistic regression classifier was used [60] to validate the performance of the top-ranked features from feature ranking. Logistic regression is commonly used for biomedical classification tasks [60,61] and in this case, can assess the association of multiple variables with an outcome e.g., DSPN or non-DSPN. Dataset was partitioned into a 70/30% ratio for train, and test set. 5-fold cross-validation was used to train the LR model. Different performance parameters were calculated to evaluate the performance of the model.

*2.5 Development and Validation of Logistic Regression-Based Nomogram*

A diagnostic nomogram was constructed by Alexander Zlotnik's [62] using multivariate logistic regression analysis in Stata/MP software (StataCorp LLC, Texas, USA). The multivariate logistic regression model was developed for two-class DSPN and Non-DSPN. The coefficients calculated from the LR model was be used to calculate linear prediction as shown in eq 1 and 2. From the linear prediction, the probability of having DSPN was calculated using eq 3.

$$coefficients = \frac{p}{1-p} \tag{1}$$

$$Linear\ Prediction\ (LP) = \ln\left(\frac{p}{1-p}\right) \qquad (2)$$

$$p = \frac{e^{LP}}{1-e^{LP}} \qquad (3)$$

The Top-ranked features (independent variables) exhibiting best performance with the LR classifier was used to create the logistic regression-based nomogram. Calibration curves were plotted to evaluate the model performance. Decision curve analysis (DCA) was carried out to identify the threshold values in which nomograms were clinically useful, using Stata software.

*2.6. Development and Validation of Severity Grading Score*

From the nomogram, a 4 class DSPN severity scoring technique was proposed based on the probability cut-off values. The performance of the proposed grading system has been validated with EDIC ground truth and the proposed grading system by Watari et al [43].

**3. Results**

*3.1. Patient's Characteristics, and Clinical Outcomes*

The EDIC patients' baseline demographic variables are shown in Table 1. The details on EDIC patients can be found in other EDIC studies [47–51]. From the collected dataset, 3754 unique data samples were retrieved after removing duplicate responses. Amon 3754 samples, 2177 samples of non-DSPN, and 1577 samples from DSPN patients. Top-10 ranked MNSI features using the extra tree feature ranking techniques has been shown in Figure 1. The Top 10 ranked features are 10-gm filament, vibration perception (R), vibration perception (L), appearance of deformities, appearance of callus, previous diabetic neuropathy, appearance of fissure, numb leg, burning leg, bed cover touch. The results of the XGboost and RF feature ranking techniques are shown in Supplementary Figures S1-S2. There is no difference found in the ranked features by the extra tree and RF technique. Therefore, for further analysis, we have studied the extra tree and Xgboost technique to find the best performing feature combination.

**Table 1.** baseline characteristics of the EDIC patients

| N: 1341<br>M: 658 (52.39%)<br>F: 598 (47.61%) | Mean | Std. Error of Mean | Minimum | Maximum |
| --- | --- | --- | --- | --- |
| AGE (years) | 35.98±6.95 | 0.19 | 20.16 | 50.99 |
| HBA1C (%) | 8.23±1.39 | 0.04 | 0.00 | 14.00 |
| BMI (Kg/m$^2$) | 26.24±4.16 | 0.11 | 0.00 | 49.82 |
| Diabetic Duration (years) | 14.55±4.91 | 0.13 | 0.00 | 27.00 |
| HDL Cholesterol (mg/dl) | 52.69±16.05 | 0.44 | 0.00 | 121.00 |
| LDL Cholesterol (mg/dl) | 110.75±36.33 | 0.99 | 0.00 | 280.00 |

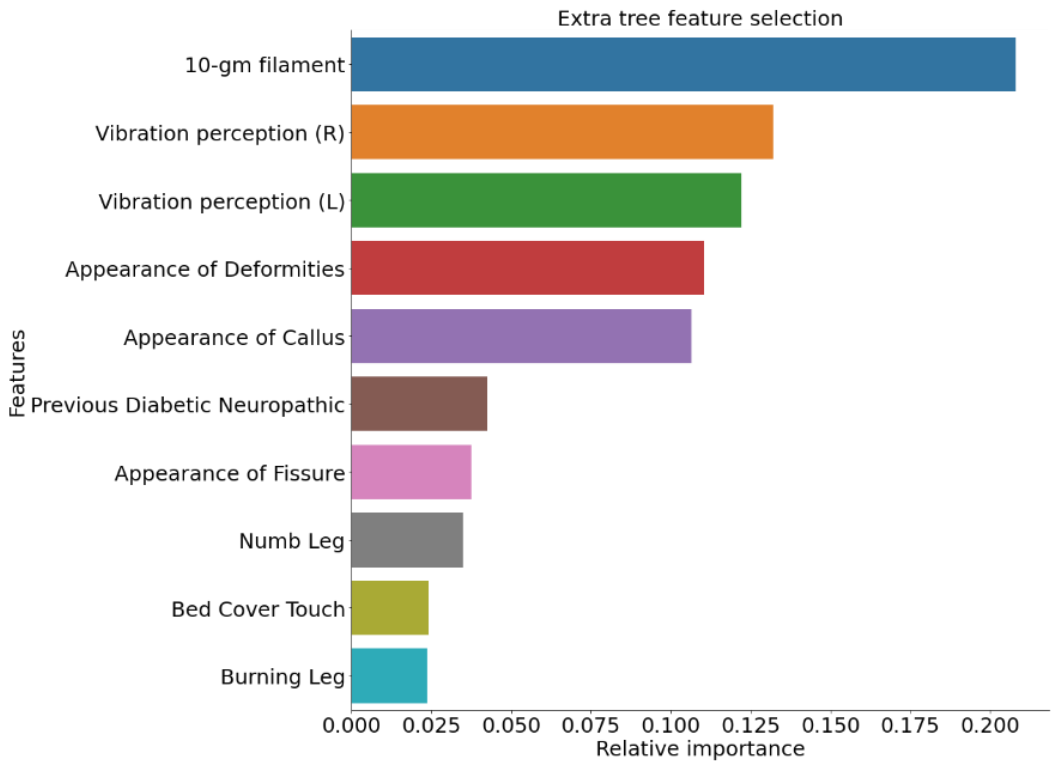

**Figure 1.** Top-ranked 10 features identified using Extra Tree algorithms from data imputed using MICE algorithm.

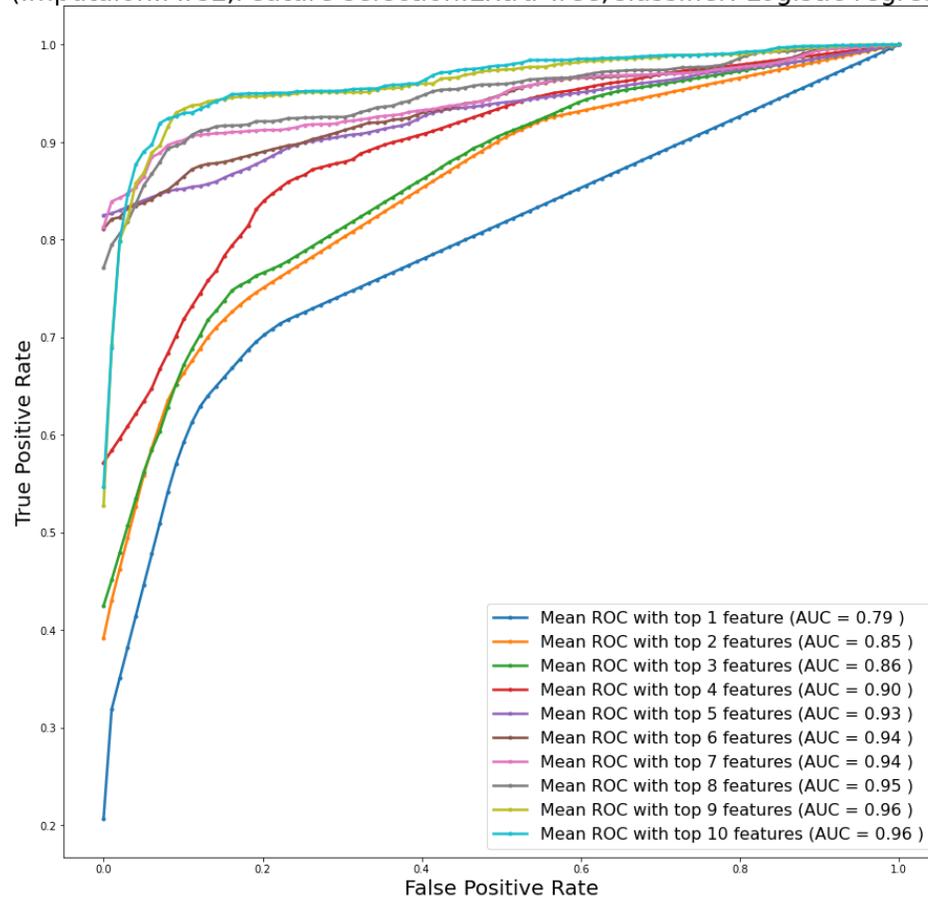

**Figure 2.** The receive operating characteristic (ROC) plots for top-ranked 1 up to 10 features using the MICE data imputation and logistic regression classification techniques for Extra Tree feature selection algorithms.

**Table 2.** Comparison of the average performance matrix and confusion matrix from five-fold cross-validation for top1 to 10 features using MICE data imputation and logistic regression classification techniques for Extra Tree feature selection algorithms.

|  | Sensitivity (%) | Specificity (%) | Accuracy (%) | Precision (%) | F1 Score (%) | Non-DSPN | | DSPN | |
|---|---|---|---|---|---|---|---|---|---|
|  |  |  |  |  |  | TN | FP | FN | TP |
| Top 1 Feature | 72±0.02 | 80±0.04 | 76±0.02 | 78±0.03 | 75±0.02 | 1732 | 445 | 619 | 1558 |
| Top 2 Features | 73±0.05 | 85±0.05 | 79±0.02 | 83±0.04 | 77±0.02 | 1845 | 332 | 591 | 1586 |
| Top 3 Features | 75±0.03 | 86±0.04 | 80±0.01 | 84±0.03 | 79±0.01 | 1869 | 308 | 548 | 1629 |
| Top 4 Features | 77±0.04 | 86±0.02 | 82±0.03 | 85±0.03 | 81±0.03 | 1877 | 300 | 496 | 1681 |
| Top 5 Features | 86±0.02 | 92±0.03 | 89±0.02 | 91±0.03 | 88±0.02 | 1994 | 183 | 315 | 1862 |
| Top 6 Features | 88±0.03 | 86±0.04 | 87±0.03 | 87±0.03 | 87±0.03 | 1879 | 298 | 260 | 1917 |
| Top 7 Features | 90±0.02 | 90±0.03 | 90±0.02 | 90±0.02 | 90±0.02 | 1954 | 223 | 220 | 1957 |
| Top 8 Features | 89±0.02 | 92±0.04 | 90±0.02 | 91±0.04 | 90±0.02 | 1995 | 182 | 238 | 1939 |
| Top 9 Features | 89±0.03 | 90±0.07 | 89±0.03 | 90±0.06 | 89±0.03 | 1949 | 228 | 233 | 1944 |
| Top 10 Features | 92±0.01 | 93±0.05 | 92±0.02 | 93±0.05 | 92±0.02 | 2019 | 158 | 185 | 1992 |
| Top 11 Features | 91±0.01 | 92±0.05 | 92±0.03 | 92±0.05 | 91±0.03 | 2001 | 176 | 194 | 1983 |
| Top 12 Features | 91±0.02 | 92±0.05 | 92±0.03 | 92±0.05 | 91±0.03 | 2003 | 174 | 195 | 1982 |
| Top 13 Features | 91±0.02 | 92±0.04 | 92±0.02 | 92±0.03 | 91±0.02 | 2012 | 165 | 204 | 1973 |
| Top 14 Features | 90±0.02 | 92±0.05 | 91±0.03 | 92±0.05 | 91±0.02 | 2007 | 170 | 210 | 1967 |
| Top 15 Features | 90±0.02 | 92±0.06 | 91±0.04 | 92±0.06 | 91±0.04 | 2008 | 169 | 209 | 1968 |

*3.2 Univariate Logistic Regression Model to Identify Variables Significantly Associated with DSPN*

From figure 2, it can be seen that both Top 9 and Top 10 features, having AUC of 0.96 for data imputed using MICE and extra tree feature ranking technique. Visually, it seems that model performance has been saturated after Top 9 features. To confirm and identify the best combination of features, we used logistic regression classifiers for performance evaluation. To determine the performance of the ranked features for DSPN identification, logistic regression classifier was trained with Top-1, Top-2, and up to Top-15 feature combination. Table 2 shows the overall accuracies and weighted average performance for the other matrices for different models using the Top 1 to 15 features for 5-fold cross-validation using the logistic regression classifier along with the confusion matrices for each case. With more than Top-10 features, there was no major change in the performance of the logistic regression classifier. The results from the LR classifier using the Top 10 features ranked for the Xgboost technique is reported in Supplementary Table S1.

The Top 10 features ranked using the extra tree technique have the best performance for the diagnosis of DSPN and non-DSPN patients compared to the Xgboost technique. From table 2, it is observed that Top 10 feature combinations are providing the best performance accuracy of 92% for DSPN identification. However, the Top 7 feature is exhibiting a reasonable performance in identifying both DSPN and non-DSPN classes with 90% sensitivity and specificity, hence, balance performance in identifying both classes. So, we have considered both Top 7 and Top 10 features models for further analysis to find out best feature combination among these two.

*3.3 Development and Evaluation of a Nomogram to Predict DSPN*

Tables 3 and 4 shows the LR models for Top 7 and Top 10 features respectively. In LR models, the z-value indicates the contribution of each variable used in the model to predict the output. It is visible from Tables 3 and 4, that all the features are statistically significant as the p values are less than 0.05. Now to choose best performing model, between the Top 10 and Top 7 features LR models, both the models were used on test set from the EDIC dataset and on an independent test set collected from Watari et al. [42]. Table 5 shows the performance evaluation metrics for both models. It is visible from Table 5 that, Top 10 features model has an accuracy of 91% on the test set from EDIC dataset however, it has an accuracy of 86% on the independent dataset collected from Watari et al. [42]. On the other hand, Top 7 features model exhibited consistent performance on the EDIC test set and better performance for the independent test set collected by Watari et al. [42] with an accuracy of 90% and 91% respectively. So, from this analysis, it is visible that, LR model with Top 7 feature combination is having reliable performance on both the dataset. So, for developing the nomogram and the severity grading system from it, Top 7 feature combinations: 10-gm filament, Vibration perception (R), Vibration perception (L), Appearance of Deformities, Appearance of Callus, Previous Diabetic Neuropathic, Appearance of Fissure, were used.

**Table 3.** The logistic regression analysis to construct the nomogram for DSPN prediction using top 7 variables by Extra Tree feature ranking technique.

| Outcome | Coef. | Std. Err. | z | P>z | [95% Conf. Interval] | |
|---|---|---|---|---|---|---|
| 10-gm filament | 2.514831 | 0.137814 | 18.25 | 0.00 | 2.24472 | 2.784941 |
| Vibration perception (R) | 2.399316 | 0.249416 | 9.62 | 0.00 | 1.91047 | 2.888162 |
| Vibration perception (L) | 1.932473 | 0.247976 | 7.79 | 0.00 | 1.446448 | 2.418498 |
| Appearance of Deformities | 2.413763 | 0.142204 | 16.97 | 0.00 | 2.135049 | 2.692477 |
| Appearance of Callus | 2.064003 | 0.13319 | 15.5 | 0.00 | 1.802955 | 2.325051 |
| Previous Diabetic Neuropathic | 1.053302 | 0.125036 | 8.42 | 0.00 | 0.808235 | 1.298369 |
| Appearance of Fissure | 2.602008 | 0.272765 | 9.54 | 0.00 | 2.067398 | 3.136619 |
| _cons | -5.31948 | 0.207402 | -25.65 | 0.00 | -5.72598 | -4.91298 |

**Table 4.** The logistic regression analysis to construct the nomogram for DSPN prediction using top 10 variables by Extra Tree feature ranking technique.

| Outcome | Coef. | Std. Err. | z | P>z | [95% Conf. Interval] | |
|---|---|---|---|---|---|---|
| 10-gm filament | 3.084504 | 0.1696 | 18.19 | 0.00 | 2.752094 | 3.416913 |
| Vibration perception (R) | 3.003988 | 0.285598 | 10.52 | 0.00 | 2.444225 | 3.56375 |
| Vibration perception (L) | 2.326558 | 0.282243 | 8.24 | 0.00 | 1.773372 | 2.879744 |
| Appearance of Deformities | 3.202711 | 0.176598 | 18.14 | 0.00 | 2.856585 | 3.548837 |
| Appearance of Callus | 2.886776 | 0.169801 | 17 | 0.00 | 2.553974 | 3.219579 |
| Previous Diabetic Neuropathic | 0.634693 | 0.140511 | 4.52 | 0.00 | 0.359297 | 0.910089 |
| Appearance of Fissure | 3.52151 | 0.309166 | 11.39 | 0.00 | 2.915556 | 4.127464 |
| Numb Leg | 0.941649 | 0.149556 | 6.3 | 0.00 | 0.648525 | 1.234772 |
| Burning Leg | 1.235312 | 0.153058 | 8.07 | 0.00 | 0.935324 | 1.535301 |
| Bed Cover Touch | 2.655393 | 0.244644 | 10.85 | 0.00 | 2.175899 | 3.134887 |
| _cons | -7.49854 | 0.306272 | -24.48 | 0.00 | -8.09883 | -6.89826 |

**Table 5.** Performance evaluation for the Top 7 and 10 features logistic regression model to construct the nomogram for DSPN prediction.

| Prediction model | Test Sets | Sensitivity (%) | Specificity (%) | Accuracy (%) | Precision (%) | F1 Score (%) | Confusion Matrix | | | |
|---|---|---|---|---|---|---|---|---|---|---|
| | | | | | | | Non-DSPN | | DSPN | |
| | | | | | | | TN | FP | FN | TP |
| | EDIC Test Set | 91 | 89 | 90 | 86 | 88 | 583 | 71 | 44 | 429 |

| Top 7 Variable model | Independent Test Set | 91 | 92 | 91 | 89 | 90 | 54 | 5 | 4 | 39 |
| Top 10 Variable model | EDIC Test Set | 91 | 92 | 91 | 89 | 90 | 598 | 56 | 42 | 431 |
| | Independent Test Set | 93 | 81 | 86 | 78 | 85 | 48 | 11 | 3 | 40 |

True positive (*TP*): True DSPN patients
True negative (*TN*): True Non-DSPN
False-positive (*FP*): Non-DSPN patients, classified as DSPN patients.
False-negative (*FN*): DSPN patients, classified as non-DSPN patients

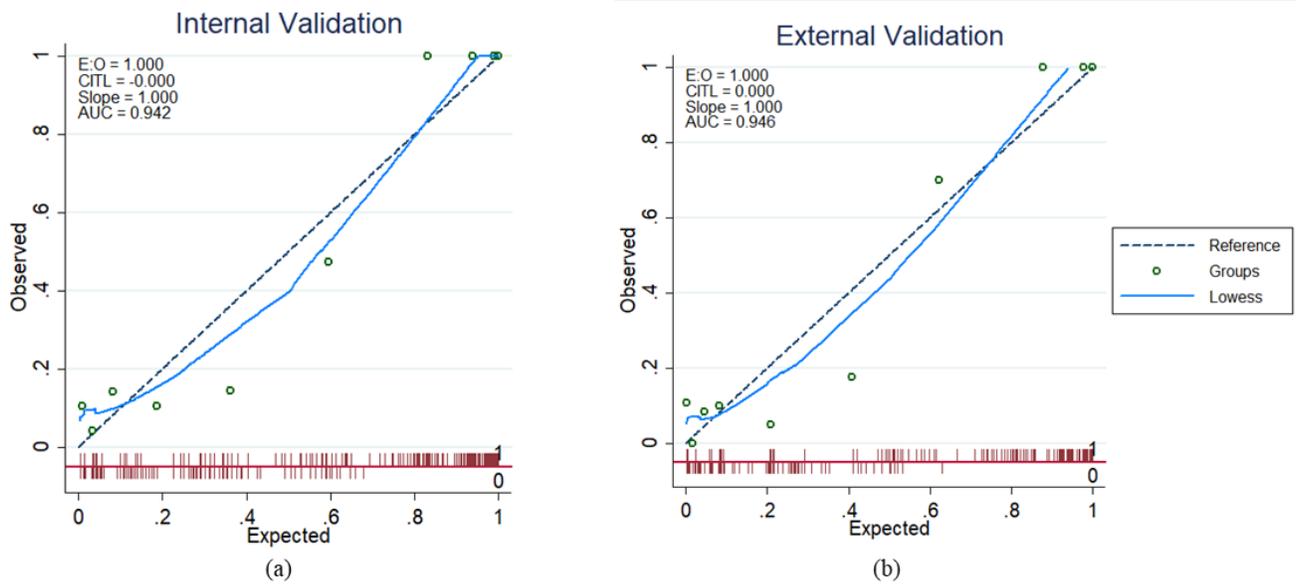

**Figure 3.** Calibration plot comparing predicted and actual DSPN probability for (a) the internal validation and (b) the external validation.

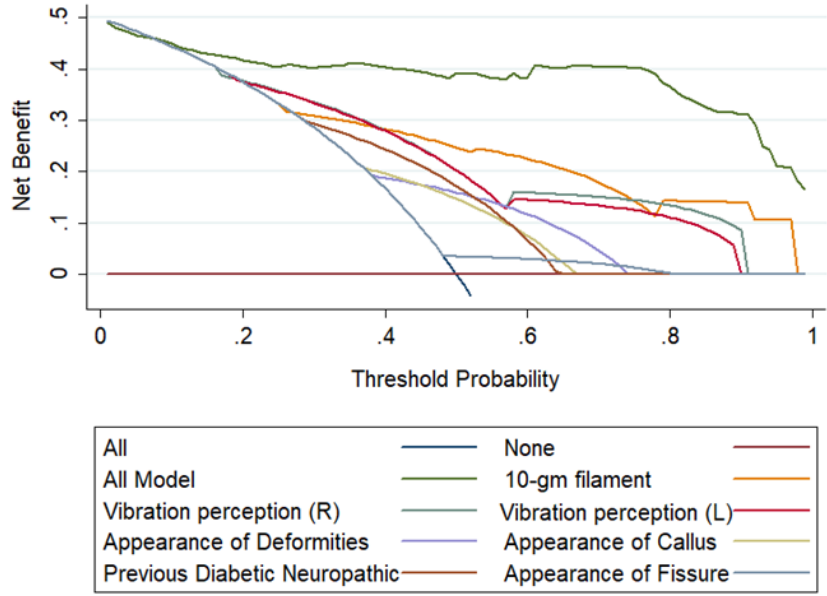

**Figure 4.** Decision curves analysis comparing different models to predict the probability of DSPN Severity. The net benefit balances the risk and potential harm from unnecessary over-intervention for patients with DSPN.

Figure 3 shows the calibration plot for train set (internal validation) and test set (external validation) with an area under the curve (AUC) of 0.94 for both, indicating good reliability of the LR model. Figure 4 illustrates the decision curve analysis comparing the net benefit of all different models created from individual features for DSPN probability prediction. It also shows the performance of the overall model (all features) net benefit for DSPN probability prediction.

Figure 5 is showing the nomogram generated using multivariate logistic regression for DSPN probability prediction using Top 7 MNSI features. The nomogram is comprised of 10 rows. Top 1-7 rows representing 7 MNSI variables along with a scale indicating the corresponding responses. The 8th row is the score scale for the responses of the seven variables. Row 9 is the probability axis indicating the probability of having DSPN of patients based on the MNSI responses. Row 10 is the Total score scale, where all the score for each MNSI response is added to calculate the final score. Figure 6 shows an example nomogram-based scoring system for a DSPN patient with the variable values at baseline. Individual scores for each predictor were calculated and added to produce the total score and the DSPN probability was calculated to 98% and according to Table 7, the patient has severe DSPN. The DSPN probability of a patient can also be calculated using eq 4 and 5, which was derived from the LR model for Top 7 features (Table 3).

$$\begin{aligned}
Linear\ prediction\ (LP) =\ & (-5.31948) \\
& + (2.514831 * 10-gm\ filaments) \\
& + (2.399316 * Vibration\ perception\ (R)) \\
& + (1.932473 * Vibration\ perception\ (L)) \\
& + (2.413763 * Appearance\ of\ Deformities) \\
& + (2.064003 * Appearance\ of\ Callus) \\
& + (1.053302 * Previous\ Diabetic\ Neuropathic) \\
& + (2.602008 * Appearance\ of\ Fissure)
\end{aligned} \quad (4)$$

$$DSPN\ Probability = \frac{1}{1 + e^{-LP}} \quad (5)$$

For each MNSI responses, a score was generated by the nomogram. Supplementary Table S2 (supplementary material) is showing the MNSI responses and their corresponding score. All the score corresponding to the MNSI responses were added together to get the total score and using the total score, DSPN probability was calculated from the nomogram. Now using the total score and corresponding probability, a 4-class severity grading system was developed as shown in Table 7. The probability values less than 50%, between 50% and 75 % and between 75% and 90%, and more than 90 % were categorized to Absent, Mild, Moderate, and Severe groups, respectively.

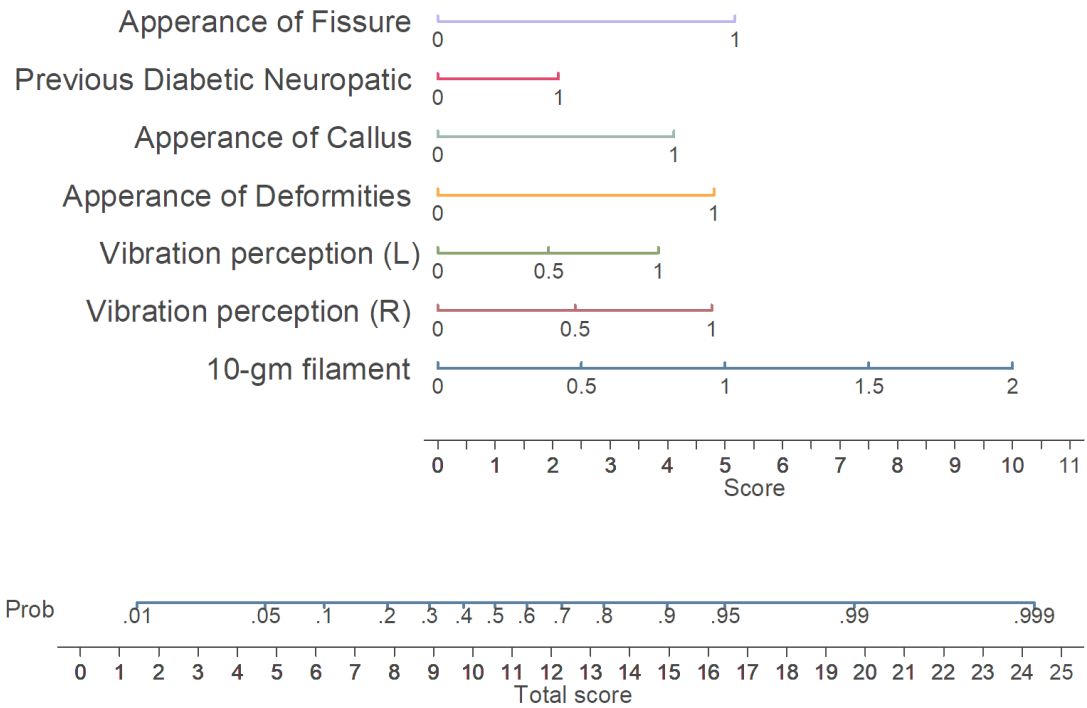

**Figure 5.** Multivariate logistic regression-based Nomogram to predict the probability of DSPN Severity. Nomogram for prediction of DSPN Severity was created using the following five predictors: 10-gm filament, Vibration perception (R), Vibration perception (L), Appearance of Deformities, Appearance of Callus, Previous Diabetic Neuropathic, Appearance of Fissure.

**Table 6.** MNSI Severity score from nomogram and corresponding severity probability of DSPN patients

| Patient Group | Absent | | | | Mild | | | | | Moderate | | | | Severe | | | |
|---|---|---|---|---|---|---|---|---|---|---|---|---|---|---|---|---|---|
| MNSI Severity score | 0 | 1 | 6.2 | 10.5 | 10.6 | 11.4 | 11.8 | 12.3 | 12.7 | 12.8 | 13.3 | 14 | 15 | 15.1 | 16.5 | 19 | >28 |
| DSPN Severity probability | 0.5 | 1 | 10 | 49 | 50 | 60 | 65 | 70 | 74 | 75 | 80 | 85 | 90 | 91 | 95 | 99 | 99.99 |

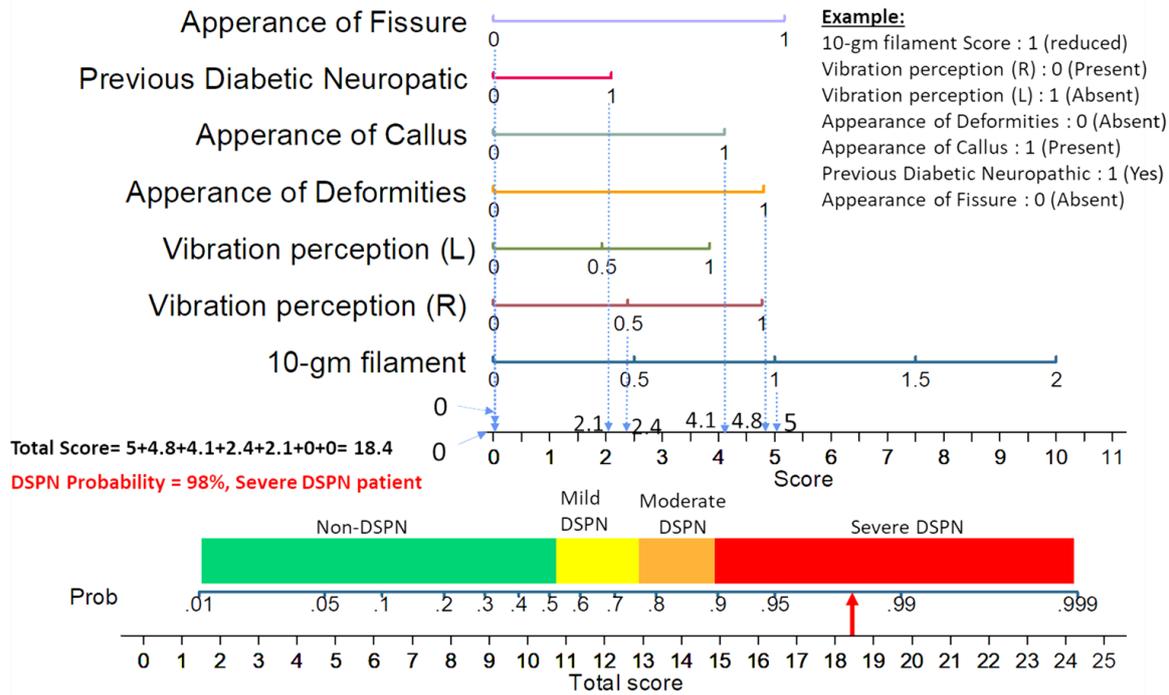

**Figure 6.** An example nomogram-based score to predict the probability of DSPN Severity.

*3.4 Performance Evaluation of the Nomogram Model*

We have applied the developed grading system on train, test and independent test set and classified patients into four DSPN severity classes (Absent/ non-DSPN, Mild, Moderate, and Severe DSPN).For EDIC train and test set, patients severity classes were cross-corelated with EDIC binary ground truth as shown in Table 7 and 8. For the EDIC training set (Table 7), out of 1526 absent class patients by the proposed grading system, 89.2% were as non-DSPN and 10.8% as DSPN patients as per EDIC ground truth; in 292 mild DSPN patients, 55.5% patients were non-DSPN and 44.5% were DSPN; for moderate and severe patients, all were DSPN patients. For the EDIC test set (Table 8), among, 635 absent class patients, 91.8% of patients were non-DSPN and 8.2% were DSPN, from 145 mild patients 48.97% were non-DSPN patients and 51.03% were DSPN patients and for Moderate and severe class all patients were DSPN based on EDIC ground truth. For the independent test set (Table 9), 93.1% and 6.9% of the patients in the absent class were non-DSPN and DSPN patients, respectively, whereas in the mild class the proportion of DSPN and non-DSPN patients were 50% and in the moderate and severe class, no patient was miss-classified. Watari et al. [42] proposed a DSPN severity grading system with the help of a Fuzzy Inference system (FIS) using three MNSI variables (questionnaire, vibration perception and 10-gm monofilament). In the MNSI dataset from Watari et al. [42], patients severity class graded using their proposed grading system was available. In Table 10, we have compared their results with our prediction models on the MNSI data set collected from study [42]. From Table 10 it can be visible that, according to Watari et al. [42] , among 102 patients, 29, 25, 27 and 21 were in absent, mild, moderate and severe class patients respectively, whereas based on our proposed model, 59, 10, 9 and 25 patients were in absent, mild, moderate and severe class patients respectively. So from Table 10, it is visible that, patients from Watari et al. [42] study, graded using the proposed grading system are not in agreement with their grading system. According to the EDIC definition for DSPN [18,51], there were 59 non-DSPN patients and 43 DSPN patients in the study by Watari et al. [42] as shown in Table 11. However, the fuzzy system classified 29 patients as non-DSPN and 73 as DSPN and as per our proposed grading system, the dataset had 58 non-DSPN and 44 DSPN patients. It is visible that, the proposed grading system agrees with EDIC definition for DSPN [18,51]. Watari et al. [42] selected only three variables (Questionnaire, vibration perception, and tactile sensitivity) as their system inputs and because the fuzzy inference system is an if/then rule-based system, there is a possibility of bias due to an inadequate number of variables for accurate identification of DSPN. Our prediction model could detect the moderate and severe class accurately without any misclassification for the training, test, and independent test cohort, with good accuracy in detecting the absent DSPN class patients (Tables 7-9).

**Table 7.** Association between different DSPN severity groups and actual outcome in the EDIC training cohort using Fisher exact probability test.

| DSPN Severity Class | Outcome | | Total |
|---|---|---|---|
| | Non-DSPN | DSPN | |
| Absent | 1,361 (89.2%) | 165 (10.8%) | 1,526 (100%) |
| Mild | 162 (55.5%) | 130 (44.5%) | 292 (100%) |
| Moderate | 0 (0%) | 282 (100%) | 282 (100%) |
| Severe | 0 (0%) | 947 (100%) | 947 (100%) |
| Total | 1,523 (50%) | 1,524 (50%) | 3,047 (100%) |

**Table 8.** Association between different DSPN severity groups and actual outcome in the EDIC testing cohort using Fisher exact probability test.

| DSPN Severity Class | Outcome | | Total |
|---|---|---|---|
| | Non-DSPN | DSPN | |
| Absent | 583 (91.8%) | 52 (8.2%) | 635 |
| Mild | 71 (48.97%) | 74 (51.03%) | 145 |
| Moderate | 0 (0%) | 120 (100%) | 120 |
| Severe | 0 (0%) | 407 (100%) | 407 |
| Total | 654 (50%) | 653 (50%) | 1,307 |

**Table 9.** Association between different DSPN severity groups and actual outcome in the independent test cohort from Watari et al., [42] using Fisher exact probability test.

| DSPN Severity Class | Outcome | | Total |
|---|---|---|---|
| | Non-DSPN | DSPN | |
| Absent | 54 (93.1%) | 4 (6.9%) | 58 (100%) |
| Mild | 5 (50%) | 5 (50%) | 10 (100%) |
| Moderate | 0 (0%) | 9 (100%) | 9 (100%) |
| Severe | 0 (0%) | 25 (100%) | 25 (100%) |
| Total | 59 (57.8%) | 43 (42.2%) | 102 (100%) |

**Table 10.** Association between different DSPN severity groups and severity grading by Watari et al. [42] in the independent test cohort using Fisher exact probability test.

| DSPN Severity Grading by our model | DSPN Severity Grading by Watari et. al. 2014 [42] | | | | Total |
|---|---|---|---|---|---|
| | Absent | Mild | Moderate | Severe | |
| Absent | 28 | 18 | 12 | 0 | 59 |
| Mild | 0 | 2 | 5 | 3 | 10 |
| Moderate | 1 | 3 | 3 | 2 | 9 |
| Severe | 0 | 2 | 7 | 16 | 25 |
| Total | 29 | 25 | 27 | 21 | 102 |

Table 11. Performance comparison of our proposed MNSI cut-offs for Binary classification on independent test cohort (Watari et al. 2014 [42])

| MNSI cut-off | Non-DSPN | DSPN |
|---|---|---|
| Feldman et al. [10] | 59 | 43 |
| Watari et. al. [42] | 29 | 73 |
| Our Prediction Model | 58 | 44 |

The difference in DSPN identification as per the EDIC definition and fuzzy model suggests a need to improve the latter. There was an association between different DSPN severity classes in the independent test set and the grading by Watari et. al. [42]. Watari et al. [42] had 29 absent, 25 mild, 27 moderate, and 21 severe patients, whereas our model predicted 59 absent, 10 mild, 9 as moderate and 25 as severe. Our nomogram-based model is more robust as it considers all the important MNSI parameters in DSPN prediction and severity grading compared to only a few parameters in the fuzzy model. This nomogram-based scoring technique can diagnose and predict the DSPN severity of patients into absent, mild, moderate, and severe (Table 6).

## 4. DISCUSSION

The ADA position statement advocates the use of symptoms, signs, and electrophysiological testing [1], whilst other guidelines have suggested the use of quantitative sensory testing and intraepidermal nerve fiber density for diagnosing DSPN [6–13,63,64]. However, the latter are expensive, require specialized personnel, and skin biopsy is painful and is not suitable for large clinical trials. For screening and clinical trials, composite screening methods that assess the sign and symptoms of DSPN are widely used [16]. The MNSI is one of the most commonly used screening technique for DSPN in clinical research [17,18,20–25] and large clinical trials like DCCT/EDIC [47,48], the Action to Control Cardiovascular Disease in Diabetes (ACCORD) [65].

The MNSI questionnaire and examination can identify clinical neuropathy but is not validated to grade the severity of DSPN like the neuropathy disability score (NDS) or neuropathy symptom score (NSS) [15,16]. Feldman et al., [17] suggested that when patients have a positive MNSI, they need to undergo the Michigan diabetic neuropathy score (MDNS), which includes clinical examination associated with nerve conduction studies (NCS). NCS is complex to perform and has large inter-individual variability both in cross-sectional and longitudinal studies and is not suitable for large clinical trials or is widely available in all healthcare facilities. A simple and reliable DPSN severity scoring system is highly desirable to identify patients with mild disease and also those at high risk for foot ulceration.

Therefore, using a state-of-the-art machine learning algorithm, a prediction scoring system using MNSI was developed and implemented to classify patients in the DCCT/EDIC clinical trial into absent, mild, moderate, and severe DSPN. The original dataset from the EDIC clinical trials had missing and duplicate responses for many patients. After removing duplicate samples, dataset was imputed using MICE algorithm to predict the missing values of the variables. MNSI variables were ranked based on their importance index for DSPN identification by different feature ranking techniques and the Extra tree algorithm was found best performing algorithm for identifying best combination of MNSI variables. Logistic regression classifier was trained for the Top 1, Top 2, up to 15 features combinations using five-fold cross-validation for identifying the best combination of features. Two models with the Top 7 and Top 10 variables showed promising results with AUC of 94% and 96% respectively. Although the Top 10 models showed better AUC, sensitivity, accuracy compared to the 7 Top-ranked features' model, when both models were validated on an external independent dataset by Watari et al. [42], the 7 Top-ranked features models performed better than the 10 Top-ranked features model and in the test set, only marginal improvements were achieved by using the Top 10 ranked features model from the Top 7 feature model. Top 7 ranked features were selected to develop the nomogram using a multivariant logistic regression model. From the nomogram, DSPN severity grading system was proposed based on the predicted DSPN probability and total score on MNSI.

In one of our previous works, [66] a nomogram based DSPN severity grading system was developed using 10 MNSI variables Appearance of Feet (R), Ankle Reflexes (R), Vibration perception (L), Vibration perception (R), Appearance of Feet (L), 10-gm filament (L), Ankle Reflexes (L), 10-gm filament (R), Bed Cover Touch, and Ulceration (R). The nomogram-based prediction model exhibited an accuracy of 97.95% and 98.84% for the EDIC test set and an independent test set collected from Diabetes Prevention Program Outcomes Study (DPPOS) clinical trials. However, in this work [66], the model performance was validated using the binary ground truth from EDIC and DPPOS. In the current study, the developed grading model was validated using an existing DSPN severity grading system, proposed by Watari et al. [42] and showed that the proposed grading model is exhibiting better then the proposed grading model by Watari et al. [42] and also in agreement with EDIC.

The strength of our study is that it was undertaken using data from a large cohort of patients from the established DCCT/EDIC trials. Our prediction model detected moderate and severe DSPN without any misclassification for the train, test, and independent test set, and exhibited good accuracy for absent DSPN. However, misclassification was evident in the mild class. The ground truth of MNSI is not 100% sensitive as it relies on large fibre damage which would miss early small fibre damage which is evident to be the mild class. The model performed well for both type 1 and type 2 diabetic patients. The study highlights the potential for machine learning-based applications to diagnose and stage the severity of DSPN. In conclusion, a machine learning-based scoring system using MNSI was developed and validated to classify the severity of DSPN. The proposed model can support clinical professionals and researchers as a secondary decision-making system.

**Funding:** This research is financially supported by Qatar National Research Foundation (QNRF) grant no. NPRP12s-0227-190164, International Research Collaboration Co-Fund (IRCC) grant: IRCC-2021-001, Universiti Kebangsaan Malaysia (UKM), Grant Number GUP-2021-019, Grant Number DIP-2020-004, and ASEAN-India Collaborative Research Project, Department of Science and Technology - Science and Engineering Research Board (DST-SERB), Govt. of India, Grant No. CRD/2020/000220. Open Access publication of this article is supported by Qatar National Library. The statements made herein are solely the responsibility of the authors.

**Acknowledgements:** We would like to thank the National Institute of Diabetes and Digestive and Kidney Diseases (NIDDK) for providing the Diabetes Control and Complications Trial / Epidemiology of Diabetes Interventions and Complications (DCCT/EDIC) database. The Diabetes Control and Complications Trial (DCCT) and its follow-up the Epidemiology of Diabetes Interventions and Complications (EDIC) study were conducted by the DCCT/EDIC Research Group and supported by National Institute of Health grants and contracts and by the General Clinical Research Center Program, NCRR. The data [and samples] from the DCCT/EDIC study was supplied by the NIDDK Central Repositories. This manuscript was not prepared under the auspices of the DCCT/EDIC and does not represent analyses or conclusions of the Research Group, the NIDDK Central Repositories, or the NIH. The database is available on request from the NIDDK website (https://repository.niddk.nih.gov/studies/edic/).

**Conflicts of Interest:** "The authors declare no conflict of interest"